\documentclass{article} 
\usepackage{iclr2027_conference,times}
\usepackage{amssymb}

\usepackage{amsmath,amsfonts,bm}

\def\eqref#1{equation~\ref{#1}}

\def\1{\bm{1}}

\DeclareMathAlphabet{\mathsfit}{\encodingdefault}{\sfdefault}{m}{sl}
\SetMathAlphabet{\mathsfit}{bold}{\encodingdefault}{\sfdefault}{bx}{n}

\usepackage{amssymb}
\usepackage{hyperref}
\usepackage{url}
\usepackage{xcolor,colortbl}
\definecolor{sparebg}{HTML}{EEF5FB}
\usepackage{booktabs}
\usepackage{multirow}
\usepackage{graphicx}
\usepackage{wrapfig}
\usepackage{needspace}
\usepackage{placeins}

\title{SPARE: Shared Perception and Adaptive Retrieval of Evidence for Streaming Video Understanding}

\author{%
Senbin Xu$^{1,*}$,\hspace{0.6em}
Zijun Zhang$^{2,*}$,\hspace{0.6em}
Xinhua Wu$^{1}$,\hspace{0.6em}
Li He$^{3}$,\hspace{0.6em}
Jubo Feng$^{4}$ \\
\bfseries
Li Qiao$^{5}$,\hspace{0.6em}
Zhen Gao$^{2}$,\hspace{0.6em}
Yan Zhang$^{1}$,\hspace{0.6em}
Jixiang Luo$^{6,\dagger}$,\hspace{0.6em}
Rongrong Ji$^{1}$ \\[0.5em]
{\normalfont\small
$^{1}$Xiamen University \quad
$^{2}$Beijing Institute of Technology \quad
$^{3}$Fudan University} \\
{\normalfont\small
$^{4}$Shanghai Jiao Tong University \quad
$^{5}$University of Hong Kong \quad
$^{6}$China Telecom} \\[0.4em]
{\normalfont\footnotesize
$^{*}$Equal contribution. \quad
$^{\dagger}$Project leader.} \\[0.2em]
{\normalfont\footnotesize
Correspondence to: \texttt{jixiangluo85@gmail.com}}
}

\iclrfinalcopy 
\begin{document}

\maketitle
\fancyhead{}
\begin{abstract}
Streaming video understanding requires answering questions from
the observed video, with evidence drawn from recent observations
or earlier events.
Historical context can improve recall but may interfere
with judgments based on recent frames.
Its usefulness depends on both when it is introduced and which
clips are selected, while relevance alone does not guarantee
sufficient evidence for an answer.
We propose \textsc{SPARE}, a framework that combines adaptive
historical retrieval with visual encoder sharing.
A lightweight gate predicts whether recent observations
provide sufficient evidence to answer the question.
When additional evidence is predicted to be necessary,
a multimodal embedding model retrieves relevant historical
clips to supplement the current context.
LoRA instruction tuning trains the understanding model to use
recent observations and retrieved history jointly and abstain
when the evidence remains insufficient.
To reduce redundant visual parameters, the understanding
and retrieval branches share a visual encoder while retaining
separate language backbones.
Four linear adapters align the shared features with the retrieval
model through distillation at the visual token and final
retrieval embedding levels.
With visual-only inputs, SPARE achieves 77.78\% on the evaluated
Real-time and Backward task groups of OVO-Bench and 73.31\%
on StreamingBench.
Compared with Qwen3-VL-8B-Instruct evaluated at 2 fps,
SPARE improves accuracy by \textbf{20.68} percentage points
on OVO-Bench and \textbf{14.15} percentage points
on StreamingBench.
\end{abstract}
\section{Introduction}
Recent multimodal large language models (MLLMs) have advanced visual
perception and video event understanding~\citep{bai2025qwen3vl,qwen35blog,an2026llavaonevision2}.
Streaming video understanding requires answering questions using only
the video observed up to the current moment.
Some questions can be answered from recent observations, whereas
others require evidence from earlier events.
Supporting both requires access to history while keeping the answer
context and its processing cost bounded.
Existing methods retain history through video KV
caches~\citep{di2025rekv,zhang2026hermes}, compressed visual
memory~\citep{zeng2025streamforest,xie2026fluxmem,fu2026streamflow},
or online-updated fast weights outside the attention context~\citep{chen2026streamttt}.

Historical context can supply evidence missing from the current view.
A pretrained MLLM using only recent frames provides a
competitive streaming baseline~\citep{shen2026simplestream}.
Historical observations can improve recall of earlier events but may
reduce the accuracy of current-scene
perception~\citep{shen2026simplestream,chen2026streamttt}.
When recent observations contain sufficient evidence,
additional history may distract the model from the information
needed to answer the question.
Video retrieval-augmented generation selects question-relevant
historical content~\citep{ataallah2024goldfish,chen2025oneclip}.
Retrieved clips may depict the objects or scenes mentioned in a
question without containing the specific event or detail required
to answer it.
The challenge is to obtain useful historical evidence while
preserving accurate judgments grounded in current observations.

\begin{figure}[t]
    \centering
    \includegraphics[width=0.90\textwidth]{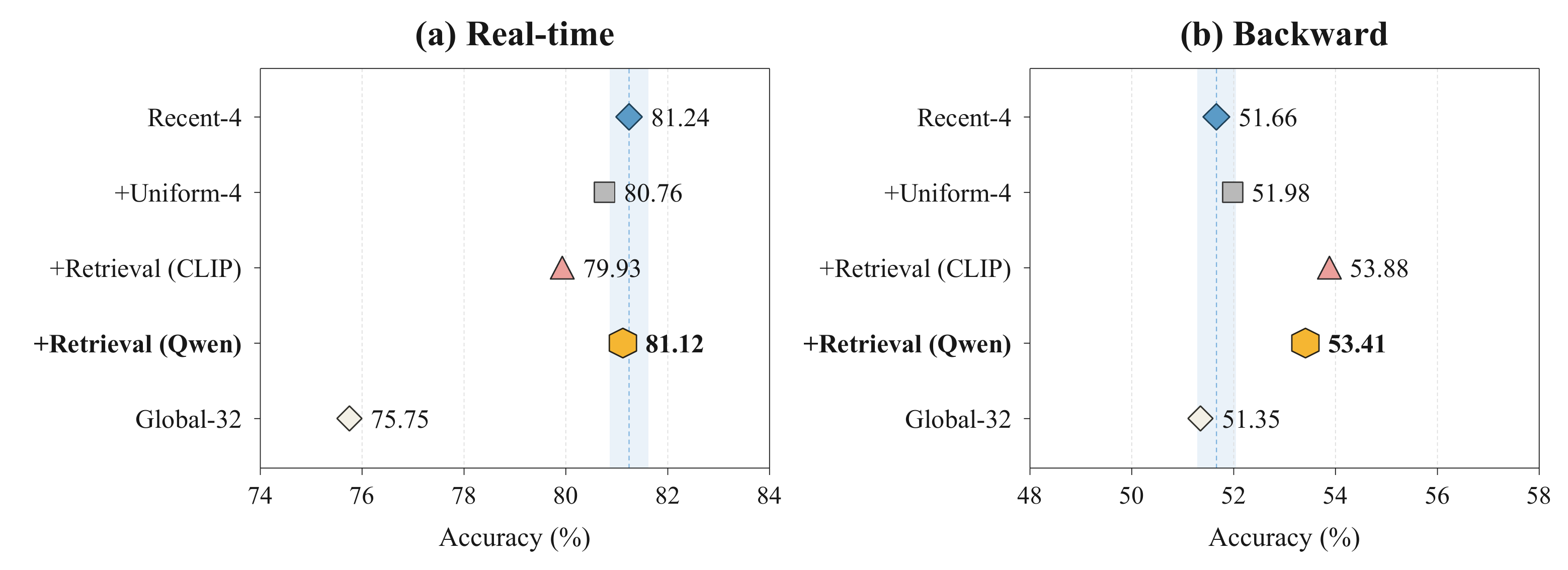}
    \caption{
Effect of historical context on OVO-Bench~\citep{li2025ovobench}.
We report question-level accuracy within
(a) Real-time (837 questions) and (b) Backward (631 questions).
All configurations use Qwen3-VL-8B-Instruct with identical
resolution and prompts, varying only input frame selection
and count.
Recent-4 uses up to four recent frames; +Uniform-4 and +Retrieval
add up to four historical frames via uniform sampling or retrieval
with CLIP ViT-L/14 or Qwen3-VL-Embedding-2B.
Both retrieval variants retrieve history for every question
without gating.
Global-32 uniformly samples up to 32 frames from the observed
video prefix.
Blue dashed lines indicate Recent-4;
bold entries highlight Qwen-based retrieval.
    }
    \label{fig:answer_context}
\end{figure}

We investigate whether historical context can improve backward
tracing while preserving real-time perception, using
Qwen3-VL-8B-Instruct with different input frame selections
(Figure~\ref{fig:answer_context}).
Adding uniformly sampled history to Recent-4 slightly improves
backward tracing but reduces real-time accuracy.
CLIP retrieval raises backward accuracy from 51.66\% to 53.88\%,
while lowering real-time accuracy from 81.24\% to 79.93\%.
With the same eight-frame budget, Qwen3-VL-Embedding-2B retrieval
achieves 53.41\% backward accuracy and 81.12\% real-time accuracy.
It yields a smaller backward gain than CLIP but largely preserves
real-time performance.

This comparison highlights the importance of retrieved clip quality:
the value of historical evidence should be assessed by both its
contribution to backward tracing and its effect on current-scene
judgments.
Global-Uniform-32 performs worse than Recent-4 on both task groups,
showing that broader coverage alone does not ensure better answers.
The experimental findings motivate considering both whether
additional history is needed and which clips can usefully
supplement recent observations.

To address these two considerations, we propose \textsc{SPARE},
a streaming video understanding framework that combines evidence
sufficiency prediction with question-conditioned historical retrieval.
A lightweight head assesses whether recent observations suffice
to answer: if so, the model answers directly; otherwise,
a multimodal embedding model retrieves historical clips by their
embedding similarity to the question to augment the current context.
LoRA fine-tuning trains the understanding model to use recent
and retrieved evidence jointly and abstain when it remains insufficient.

To reduce redundant visual parameters and encoding computation,
the two branches share the understanding model's visual encoder
while retaining separate language backbones.
Four linear adapters map shared visual features into the retrieval
model's input space, removing its standalone visual tower.
The adapters are distilled from a pretrained multimodal embedding
model with the same backbone architecture as the understanding model,
using supervision on both visual-token features and final
retrieval embeddings.

We evaluate SPARE on OVO-Bench~\citep{li2025ovobench} and
StreamingBench~\citep{lin2026streamingbench} with visual-only inputs.
SPARE achieves 77.78\% on the evaluated Real-time and Backward
tasks of OVO-Bench and 73.31\% on StreamingBench.
As shown in Table~\ref{tab:main_results}, SPARE outperforms
Qwen3-VL-8B-Instruct evaluated at 2 fps by 20.68 percentage
points on OVO-Bench and 14.15 on StreamingBench.

Ablations show that on-demand retrieval substantially reduces history use while retaining nearly the same accuracy as always retrieving.
With the same maximum number of input frames, combining recent
frames with retrieved history outperforms using recent frames alone.
Efficiency analyses show a mean per-question peak answering
context of $2.87\mathrm{K}$ tokens and stable response latency
across the tested observed frame counts.

Our main contributions are summarized as follows:
\begin{itemize}
    \item We propose \textsc{SPARE}, a streaming video understanding
    framework that predicts whether recent observations provide
    sufficient evidence and retrieves history based on the question
    when needed. LoRA fine-tuning trains the understanding model to use
    available evidence and abstain when it remains insufficient.

    \item We share the visual encoder between understanding and
    retrieval through four linear adapters trained with distillation
    on visual token features and final retrieval embeddings.
    This removes the retrieval branch's standalone visual encoder
    while retaining both language backbones.

    \item We demonstrate improvements over native Instruct models
    on OVO-Bench and StreamingBench using only visual inputs.
    Ablations examine the effects of encoder sharing,
    LoRA tuning, and input frame allocation.
\end{itemize}

\section{Related Works}
\label{sec:related_works}
\subsection{Streaming Video Understanding}
\label{sec:streaming_video_understanding}
Offline video understanding assumes a fully observed sequence, whereas
streaming methods maintain access to past observations as new frames
arrive. They use visual-token pruning, KV-state selection, recurrent
encoding, or hierarchical memory~\citep{yao2025timechatonline,di2025rekv,qian2024streaming,zeng2025streamforest,xie2026fluxmem};
other systems decouple continuous perception from response
generation~\citep{qian2025dispider}. Recent-window baselines show that
history can improve recall while impairing current-scene
perception~\citep{shen2026simplestream}, and post-training can improve
performance even under a fixed recent-window protocol~\citep{wu2026streamopd}.

\subsection{Image--Text Matching and Multimodal Embeddings}
\label{sec:embedding_models}
CLIP learns comparable image--text embeddings through dual encoders and
symmetric InfoNCE, using paired examples as positives and other batch
elements as negatives~\citep{radford2021learning}. SigLIP instead applies
independent sigmoid losses, avoiding global batch
normalization~\citep{zhai2023sigmoid}. Multimodal embedding models extend
this approach to instruction-conditioned retrieval over text, natural
and document images, and video~\citep{jiang2025vlm2vec,li2026qwen3vlembedding}.
Multi-task contrastive learning, hard negatives, and reranker distillation
adapt similarity to different relevance criteria; related work also
studies video-conditioned answerability~\citep{yoon2025can}.
We use pretrained multimodal embeddings to index historical clips
once and rank them for later queries, with visual alignment adapting
the shared encoder to the retrieval space.
\section{Method}
\label{sec:method}
\subsection{Overview} 
\label{sec:overview}
\begin{figure}[t]
    \centering
    \includegraphics[width=\linewidth]{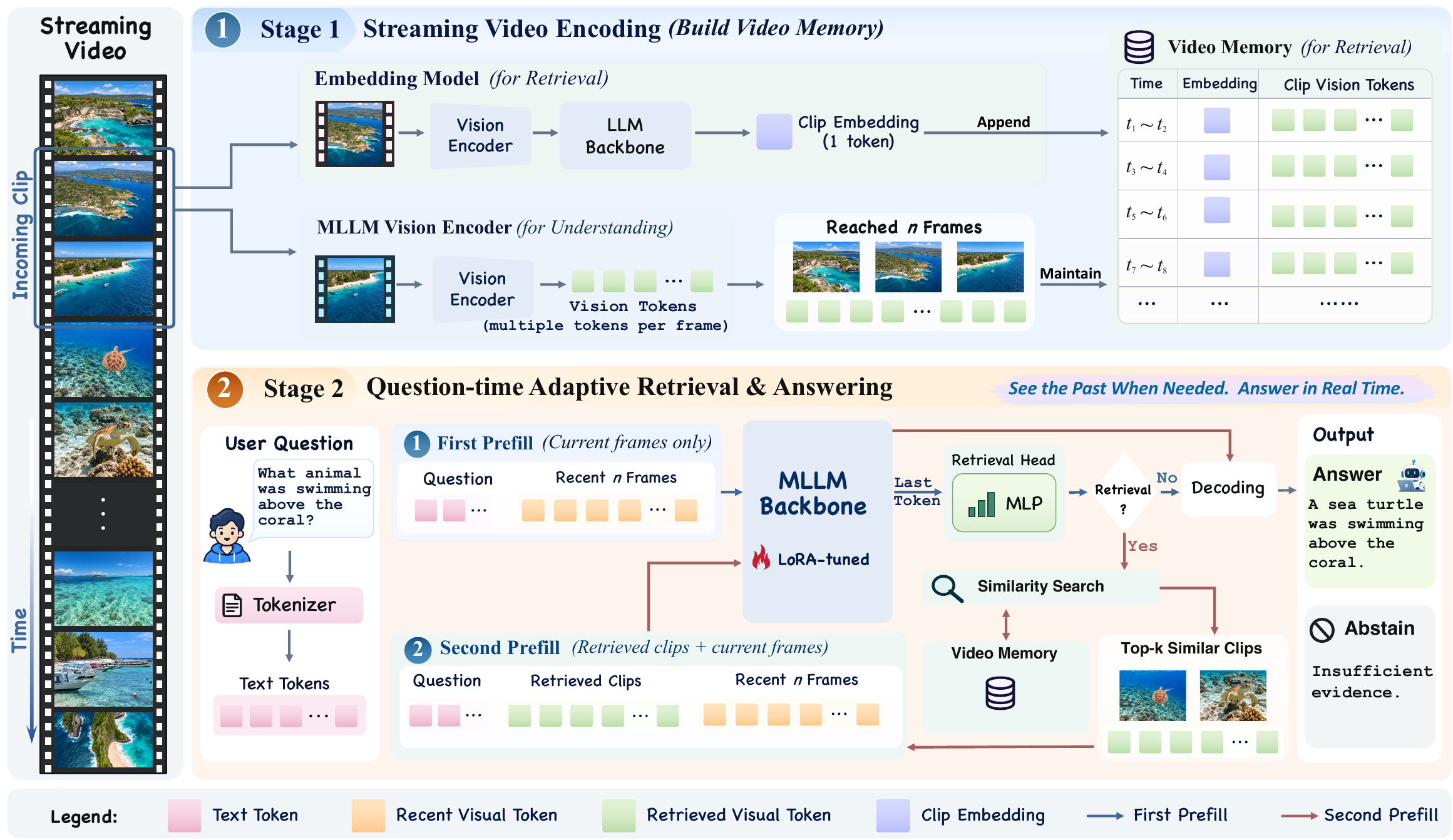}
\caption{Overview of \textsc{SPARE}.
    \textbf{Memory construction:} the two branches share visual-encoder
    weights and index historical clips by embeddings and timestamps.
    \textbf{Adaptive answering:} a gate reads the current-input prefill
    state and triggers historical retrieval when evidence is insufficient.
    The retrieved clips and current window form a fresh second-pass input;
    otherwise, the first-pass answer is retained.
    LoRA tuning supports answering or abstention from the supplied evidence.}
    \label{fig:spare_pipeline}
\end{figure}

SPARE combines video understanding with historical retrieval to answer questions about an observed video stream.
Figure~\ref{fig:spare_pipeline} summarizes the memory construction and adaptive answering pipeline.
At query time $t$, SPARE answers $q_t$ from the observed video prefix $\mathcal{V}_{\leq t}$.
The current context $\mathcal{C}_t$ contains the most recent $n$ frames, while earlier clips populate an incrementally updated visual memory $\mathcal{M}_t$.
The MLP retrieval gate (labeled Retrieval Head in Figure~\ref{fig:spare_pipeline}) assesses whether the current context provides sufficient evidence to answer the question.
If the evidence is insufficient and history is available, we construct a fresh input with the retrieved historical clips preceding the current frames, preserving their temporal order for a second answering pass.
The same LoRA adapter in the understanding model is enabled in both passes, allowing the model to answer or abstain based on the supplied evidence.

The understanding and retrieval components can operate independently with their native visual encoders and language backbones.
SPARE connects them through the clips selected from memory and optionally shares their visual encoder to remove the parameter overhead of a second tower.
Figure~\ref{fig:shared_vit} contrasts the native and shared configurations.
In the shared configuration, each model retains its own language backbone and runs the shared encoder on its own inputs.
Lightweight adapters map the shared visual features into the retrieval model's input space.
We describe the visual memory, retrieval decision, and answer supervision below.
\subsection{Shared-Encoder Visual Memory Retrieval} 
\label{sec:visual_memory_retrieval}
To select historical clips more relevant to the question,
we adopt Qwen3-VL-Embedding-2B~\citep{li2026qwen3vlembedding}
as the retrieval model in our system, based on its stronger
video retrieval performance
(Appendix~\ref{app:retriever_comparison}).

\textbf{Clip-level visual memory.}
Qwen3-VL-Embedding maps historical video clips and textual queries into a common retrieval space. We divide the stream into non-overlapping $\Delta$-second clips
$X_i=\{x_{i,1},\ldots,x_{i,m}\}$ with intervals $\tau_i=[b_i,e_i)$,
sampled at $f_s=1$ FPS with $m=\Delta f_s$ frames.
Each completed clip is encoded as
\begin{equation}
    \mathbf{z}_i = \mathcal{N}\big(E_{\mathrm{vis}}(X_i)\big),
    \qquad
    \mathcal{N}(\mathbf{x}) =
    \frac{\mathbf{x}}{\|\mathbf{x}\|_2}.
    \label{eq:segment_embedding}
\end{equation}
where $E_{\mathrm{vis}}$ is the complete clip-to-embedding path and
$\mathbf{z}_i\in\mathbb{R}^{d}$.
For current-context start time $s_t$, the eligible memory is
\begin{equation}
    \mathcal{M}_t =
    \left\{
        (\mathbf{z}_i,\tau_i,\pi_i)
        \;\middle|\;
        e_i\leq s_t
    \right\},
    \label{eq:visual_memory}
\end{equation}
where $\pi_i$ indexes the original visual clip, separating search
vectors from answering inputs and excluding overlap with the current window.
The text path $E_{\mathrm{text}}$ ranks eligible clips by cosine similarity:
\begin{equation}
    \begin{aligned}
        \mathbf{z}_{q_t}
        &= \mathcal{N}\big(E_{\mathrm{text}}(q_t)\big), \\
        s_{t,i}
        &= \mathbf{z}_{q_t}^{\top}\mathbf{z}_i, \\
        \mathcal{I}_t
        &= \operatorname{TopK}_{\,i:e_i\leq s_t}(s_{t,i}).
    \end{aligned}
    \label{eq:memory_retrieval}
\end{equation}
$\operatorname{TopK}$ selects at most $K$ clips, returning all if
fewer are available. Their visual content and timestamps form
$\widehat{\mathcal{R}}_t=\{(X_i,\tau_i)\mid i\in\mathcal{I}_t\}$,
ordered chronologically for answer generation.

\begin{figure}[t]
    \centering
    \includegraphics[width=0.95\linewidth]{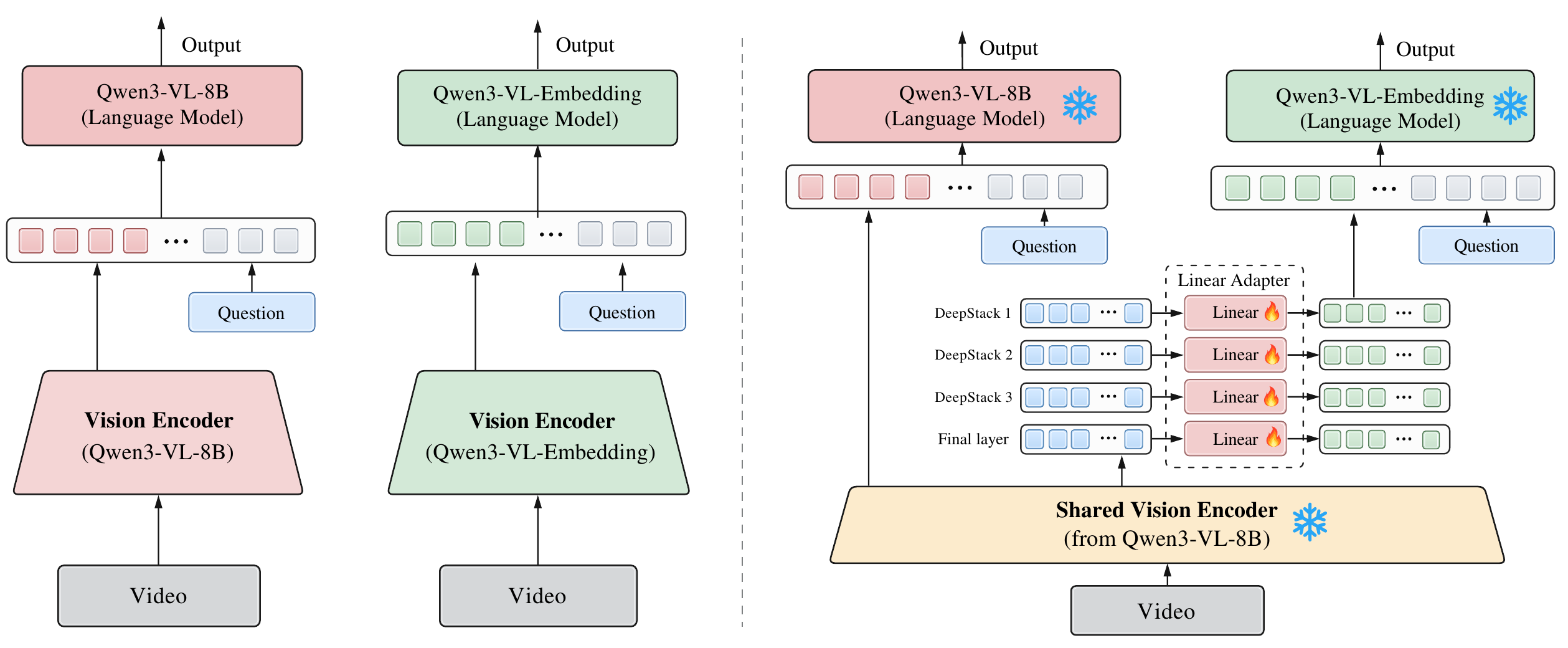}
    \caption{Shared visual encoding for understanding and retrieval.
    Left: separate visual encoders serve the two language models.
    Right: SPARE reuses the understanding model's visual encoder and
    adapts its three DeepStack merger features and final merger features through four linear adapters for the retrieval branch.
    Each branch retains its own language backbone.}
    \label{fig:shared_vit}
\end{figure}

\textbf{Shared visual encoder.}
To remove the retrieval branch's standalone visual tower, we reuse the visual encoder of the Instruct model, as described in Figure~\ref{fig:shared_vit}.
Its features must be mapped into the representation space expected by
the retrieval language model. We learn this mapping by distilling the
original Embedding-2B model into a student with the shared encoder.
Four independent linear adapters read the features immediately before
the shared mergers' output projections: three DeepStack branches and
the final branch. They map these features into the retrieval model's
input space while preserving the understanding branch's original projections.
Let $\mathbf{H}_j$ be the shared merger features,
$\mathbf{T}_j$ the corresponding teacher outputs, and $A_{\eta,j}$ the
adapters, with $j=0,1,2$ denoting DeepStack and $j=3$ the final branch.
The student features and semantic embeddings are
\begin{equation}
    \begin{aligned}
        \mathbf{S}_j
        &= A_{\eta,j}(\mathbf{H}_j),
        \qquad j\in\{0,1,2,3\}, \\
        \mathbf{e}_s
        &= G_{\psi}(\mathbf{S}_0,\mathbf{S}_1,\mathbf{S}_2,\mathbf{S}_3), \\
        \mathbf{e}_t
        &= G_{\psi}(\mathbf{T}_0,\mathbf{T}_1,\mathbf{T}_2,\mathbf{T}_3).
    \end{aligned}
    \label{eq:shared_encoder_alignment}
\end{equation}
Here, $G_{\psi}$ is the frozen retrieval language model with its original
representation extraction procedure; both paths use identical text
inputs and visual injection schemes.
To preserve the pretrained retrieval representation, we align both the visual features supplied to $G_{\psi}$ and the final clip embeddings:
\begin{equation}
    \begin{aligned}
        \mathcal{L}_{\mathrm{distill}}
        = \sum_{j=0}^{3} w_j
        \Big[
            d_{\cos}\big(\mathbf{S}_j,\operatorname{sg}(\mathbf{T}_j)\big)
            + 0.1\,\operatorname{SmoothL1}
            \big(\mathbf{S}_j,\operatorname{sg}(\mathbf{T}_j)\big)
        \Big]
        + d_{\cos}\big(\mathbf{e}_s,\operatorname{sg}(\mathbf{e}_t)\big).
    \end{aligned}
    \label{eq:distillation_loss}
\end{equation}
Here, $w_j$ weights the visual alignment loss for branch $j$;
we set $(w_0,w_1,w_2,w_3)=(0.1,0.2,0.3,0.4)$ in our experiments.
The distance $d_{\cos}$ averages $1-\cos(\cdot,\cdot)$ between
corresponding vectors over tokens for visual features and over
samples for final embeddings.
$\operatorname{SmoothL1}$ averages over all elements, and
$\operatorname{sg}$ stops gradients. Only the four adapters are updated.
At inference, the student implements $E_{\mathrm{vis}}$ in Eq.~\eqref{eq:segment_embedding}: the shared encoder, four adapters, and $G_{\psi}$ produce $\mathbf{e}_s(X_i)$, and $\mathbf{z}_i=\mathcal{N}(\mathbf{e}_s(X_i))$ is stored in memory.
The teacher's visual tower is removed; the text path and both language backbones remain unchanged.
\subsection{On-Demand Retrieval for Streaming Video Understanding}
\label{sec:on_demand_retrieval}
At query time, the model first processes the question and current frames.
The retrieval gate uses the prefill state to estimate whether this current
evidence is insufficient, before any answer tokens are generated:
\begin{equation}
    \mathbf{h}_t =
    \big[
        F_{\theta,\omega}^{(L)}
        \big(\mathcal{P}(q_t,\mathcal{C}_t,\varnothing)\big)
    \big]_{\mathrm{last}},
    \qquad
    p_t = \sigma\big(g_{\gamma}(\mathbf{h}_t)\big).
    \label{eq:gating_head}
\end{equation}
Here, $\mathcal{P}$ constructs the prompt, and $F_{\theta,\omega}$ is the $L$-layer understanding model with frozen base parameters $\theta$ and LoRA parameters $\omega$.
The state $\mathbf{h}_t$ is taken from the last valid input token after the final normalization.
The retrieval gate $g_{\gamma}$ is a two-layer MLP with layer normalization, GELU, and training-time dropout; $\sigma$ maps its logit to a probability.
Its input reflects the question and current frames, without retrieved content or generated answers.
With a default threshold of 0.5, retrieval is triggered when
$p_t\geq 0.5$ and historical memory is nonempty:
\begin{equation}
    r_t =
    \mathbb{I}[p_t\geq 0.5]\,
    \mathbb{I}[|\mathcal{M}_t|>0].
    \label{eq:retrieval_decision}
\end{equation}
$\mathbb{I}[\cdot]$ is the indicator function.
This decision selects the context for the final response, as shown in
Figure~\ref{fig:spare_pipeline}.
If $r_t=0$, the model returns its response from the current window,
with $\mathcal{R}_t=\varnothing$.
If $r_t=1$, retrieval supplies $\mathcal{R}_t=\widehat{\mathcal{R}}_t$,
and a second prefill processes the retrieved history followed by
current frames, retaining timestamps. The response from this context becomes the final output.

The same LoRA adapter supports answering or abstention in both passes
(Section~\ref{sec:lora_instruction_tuning}). The gate decides whether
to seek history; the understanding model decides whether the supplied
evidence supports an answer, since retrieval may still leave the
question unanswerable.

To train the gate, we label whether the current frames alone support an answer.
For this current-context answerability $a_t^{\mathrm{cur}}\in\{0,1\}$, the target is $r_t^*=1-a_t^{\mathrm{cur}}$.
We freeze the understanding model and optimize the gate on cached states with unweighted binary cross-entropy:
\begin{equation}
    \mathcal{L}_{\mathrm{gate}} =
    -\mathbb{E}_{(\mathbf{h}_t,r_t^{*})\sim\mathcal{D}_{\mathrm{gate}}}
    \big[
        r_t^{*}\log p_t
        +(1-r_t^{*})\log(1-p_t)
    \big].
    \label{eq:gate_loss}
\end{equation}
Here, $\mathcal{D}_{\mathrm{gate}}$ contains cached states and
current-evidence labels; the loss is implemented with logits.
We train the gate on original-model states, then adapt it to
LoRA-enabled states using the same labels and objective, so its inputs
match the answering model at deployment.
Appendices~\ref{app:initial_gate_training} and~\ref{app:answer_tuning_details}
provide label construction and training details.

\subsection{LoRA Instruction Tuning}
\label{sec:lora_instruction_tuning}
LoRA instruction tuning~\citep{hu2022lora} teaches the model to answer
or abstain based on combined current and retrieved evidence,
whereas the gate assesses current frames alone.
Sources cover \emph{current-answerable} questions, \emph{history-assisted}
questions requiring earlier evidence, and \emph{insufficient-evidence}
questions unsupported by the supplied current and historical observations.
We pair each question with one clip retrieved offline and the current
window, in chronological order with timestamps, sampling each at
1 FPS with up to four frames.
We reassess answerability on this combined input because retrieval
can change the source label. Sampling and consistency filtering
retain 173 examples (Appendix~\ref{app:answer_tuning_details}).

For example $i$,
$x_i=\mathcal{P}(q_i,\mathcal{C}_i,\mathcal{R}_i,\mathcal{O}_i)$
contains the question, timestamped frames, instructions, and options
$\mathcal{O}_i$ when present. The combined-evidence label $a_i$ is 1
for answerable inputs and 0 otherwise, selecting the target:
\begin{equation}
    y_i^* =
    \begin{cases}
        b_i^{\mathrm{ans}}\oplus e, & a_i=1,\\
        b_i^{\mathrm{abs}}\oplus e, & a_i=0,
    \end{cases}
    \label{eq:sft_target}
\end{equation}
where $e=\texttt{<|im\_end|>}$ and $\oplus$ denotes concatenation.
For multiple-choice questions, $b_i^{\mathrm{ans}}$ is the supported option letter and $b_i^{\mathrm{abs}}$ is the existing abstention-option letter.
For open-ended questions, they are the reviewed short answer and \texttt{Unable to answer.}, respectively.
The answerability label determines the target but is absent from the input.

These input--target pairs form $\mathcal{D}_{\mathrm{LoRA}}=\{(x_i,y_i^*)\}_{i=1}^{N}$, on which we minimize the mean answer-only loss:
\begin{equation}
    \begin{aligned}
        \mathcal{L}_{\mathrm{LoRA}}(\omega)
        &= \frac{1}{|\mathcal{D}_{\mathrm{LoRA}}|}
        \sum_{(x_i,y_i^*)\in\mathcal{D}_{\mathrm{LoRA}}}\ell_i(\omega), \\
        \ell_i(\omega)
        &= -\frac{1}{T_i}\sum_{j=1}^{T_i}
        \log p_{\theta,\omega}
        \big(y_{i,j}^*\mid x_i,y_{i,<j}^*\big).
    \end{aligned}
    \label{eq:sft_loss}
\end{equation}
Here, $T_i$ includes the end-of-turn token. Teacher forcing uses
full-vocabulary probabilities, equal weight per example, and no loss
on prompt, visual, or padding positions.
We update only language-model LoRA parameters $\omega$ for one epoch,
freezing the base parameters $\theta$, visual encoder, retriever,
and gate. We then freeze LoRA and adapt the gate to LoRA-enabled
first-pass states; both answering passes use the same adapter.
Appendix~\ref{app:answer_tuning_details} provides prompts, masking,
optimization, and gate-adaptation details.

\begin{table}[!t]
\setlength{\abovecaptionskip}{0pt}
\caption{Measured results on OVO-Bench and StreamingBench (\%).
For SPARE, $a+b$ denotes the frame counts in the first and retrieval-triggered second answering passes.
Native and shared ViT refer to the retrieval branch's visual encoder.
Bold denotes the best result and underlining the second-best result in each column, including ties.}
\label{tab:main_results}
\centering
\small
\setlength{\tabcolsep}{2pt}
\setlength{\arrayrulewidth}{0.3pt}
\renewcommand{\arraystretch}{1.12}

\resizebox{\linewidth}{!}{%
\begin{tabular}{@{}lcrrrrrrrrrrrrrrrr@{}}
\toprule
\multirow{3}{*}{Model} & \multirow{3}{*}{\shortstack{Frames /\\sampling}} & \multicolumn{4}{c}{\textit{StreamingBench}} & \multicolumn{12}
{c}{\textit{OVO-Bench}} \\
\cmidrule(lr){3-6}\cmidrule(lr){7-18}
& & \multirow{2}{*}{RTVU} & \multirow{2}{*}{OSU} & \multirow{2}{*}{CU} & \multirow{2}{*}{All} & \multicolumn{7}{c}{\textit{Real-Time Visual
Perception}} & \multicolumn{4}{c}{\textit{Backward Tracing}} & \multirow{2}{*}{All} \\
\cmidrule(lr){7-13}\cmidrule(lr){14-17}
& & & & & & OCR & ACR & ATR & STU & FPD & OJR & Avg. & EPM & ASI & HLD & Avg. & \\
\midrule

\multicolumn{18}{c}{\textit{Offline Video LLMs}} \\
\midrule
LLaVA-OneVision2-8B & 16 & 74.20 & 37.90 & 44.80 & 59.60 & 72.48 & 61.47 & 71.55 & 63.48 & 65.35 & 60.87 & 65.87 & 56.57 & 64.86 & 11.29 & 44.24 & 55.05 \\
InternVL2-8B & 32 / $\leq$64 & 67.60 & 35.20 & 36.90 & 53.58 & 69.80 & 59.63 & 67.24 & 50.56 & 71.29 & 57.61 & 62.69 & 52.53 & 58.11 & 25.81 & 45.48 & 54.08 \\
Dispider-7B (offline) & 16/clip & 68.72 & 42.20 & 29.90 & 54.20 & 57.05 & 49.54 & 62.07 & 44.94 & 60.40 & 51.63 & 54.27 & 48.82 & 55.41 & 4.30 & 36.18 & 45.22 \\

\midrule
\multicolumn{18}{c}{\textit{Online / Streaming Video LLMs}} \\
\midrule
Flash-VStream-7B & 1~fps & 53.16 & 29.40 & 23.90 & 41.38 & 41.61 & 53.21 & 48.28 & 29.21 & 60.40 & 42.93 & 45.94 & 41.75 & 41.22 & 1.08 & 28.01 & 36.98 \\
TimeChat-Online-7B & 1/0.5/0.2 fps & 73.92 & 39.70 & 32.40 & 57.09 & 73.83 & 47.71 & 69.83 & 46.63 & 71.29 & 60.33 & 61.60 & 56.90 & 64.19 & 6.99 & 42.69 & 52.15 \\
StreamForest-7B & 1~fps & 77.52 & 38.20 & 37.30 & 59.84 & 73.15 & 53.21 & 72.41 & 52.25 & 63.37 & 61.96 & 62.72 & 58.59 & 66.22 & 29.03 & 51.28 & 57.00 \\
HERMES-7B & 1~fps & 78.88 & 49.50 & 44.20 & 64.64 & 88.59 & 66.97 & 80.17 & 54.49 & 72.28 & 71.74 & 72.37 & 47.14 & 63.51 & 39.25 & 49.97 & 61.17 \\
StreamTTT-4B & 2~fps & 81.28 & 55.70 & 46.70 & 67.91 & 91.28 & 80.73 & 80.17 & 56.18 & \textbf{82.18} & 77.72 & 78.04 & \textbf{60.94} & 58.11 & \underline{60.75} & \underline{59.93} & 68.99 \\

\midrule
\multicolumn{18}{c}{\textit{Open-Source Video LLMs}} \\
\midrule
Qwen3-VL-8B-Instruct & 2~fps & 74.84 & 41.00 & 38.10 & 59.16 & 81.21 & 62.39 & 75.00 & 58.99 & 71.29 & 63.04 & 68.65 & 56.57 & \underline{68.24} & 11.83 & 45.55 & 57.10 \\
+ \textsc{SimpleStream} & 4 & 80.68 & 62.80 & 58.50 & 71.78 & 93.29 & 83.49 & 81.90 & 66.29 & \underline{78.22} & \textbf{84.24} & 81.24 & 52.53 & 56.76 & 52.69 & 53.99 & 67.61 \\

\rowcolor{sparebg}
+ \textsc{SPARE} w/ native ViT & 4+8 & \underline{83.08} & \textbf{64.50} & \textbf{62.70} & \textbf{74.42} & \underline{95.97} & \underline{84.40} & \textbf{86.21} & \underline{70.22} & 74.26 & \underline{83.15} & \textbf{82.37} & 58.92 & 66.89 & 43.01 & 56.28 & 69.32 \\
\rowcolor{sparebg}
+ \textsc{SPARE} w/ shared ViT & 4+8 & \textbf{83.20} & \underline{63.80} & 62.20 & \underline{74.22} & \textbf{96.64} & \textbf{85.32} & \underline{85.34} & \underline{70.22} & 71.29 & 82.61 & 81.91 & 59.93 & \underline{68.24} & 42.47 & 56.88 & \underline{69.39} \\
\rowcolor{sparebg}
+ \textsc{SPARE} w/ shared ViT \& LoRA & 4+8 & 82.60 & 61.10 & \underline{62.30} & 73.31 & \underline{95.97} & \textbf{85.32} & 83.62 & \textbf{70.79} & 73.27 & 82.61 & \underline{81.93} & \underline{60.27} & \textbf{70.27} & \textbf{90.32} & \textbf{73.62} & \textbf{77.78} \\
\bottomrule
\end{tabular}%
}
\end{table}
\section{Experiments}
\label{sec:experiments}
\subsection{Training Settings}
\label{sec:training_settings}
We use Qwen3-VL-8B-Instruct for understanding and Embedding-2B for retrieval.
Four visual adapters are aligned on 64K image and 64K video entries
with both backbones frozen.
The initial retrieval gate is trained on cached states with independent
current-evidence assessments and mixed question formats.
LoRA instruction tuning then updates the understanding model for one epoch on 173 audited examples
(78 multiple-choice and 95 open-ended), paired with retrieved history.
Finally, we re-extract LoRA-enabled states and adapt the gate for one
epoch on 15k audited questions using unweighted binary cross-entropy. The source videos used for gate training and adaptation
and LoRA instruction tuning do not overlap with those
used for evaluation in OVO-Bench or StreamingBench.
Appendices~\ref{app:answer_tuning_details}
and~\ref{app:alignment_gate_training} provide data construction,
prompts, and optimization settings.

\subsection{Evaluation Settings}
\label{sec:evaluation_settings}
\textbf{Benchmarks.}
We evaluate streaming video understanding on
OVO-Bench~\citep{li2025ovobench} and
StreamingBench~\citep{lin2026streamingbench} with visual-only inputs.
For OVO-Bench, we evaluate 1,468 questions covering Real-time
Visual Perception and Backward Tracing, excluding Forward Responding.
These two groups assess understanding of current observations
and recall of earlier events.
We compute the macro-average accuracy across tasks within each
group and average the two group scores for the overall result.
For StreamingBench, we evaluate all 4,500 questions from 900 videos,
covering real-time visual, omni-source, and contextual understanding.
We report accuracy for each category and compute the overall
score by weighting the three category accuracies by question count.
Appendix~\ref{app:evaluation_details} provides further dataset details.

\textbf{Compared models.}
Table~\ref{tab:main_results} includes offline
models~\citep{bai2025qwen3vl,an2026llavaonevision2,chen2024expanding},
Dispider in offline mode~\citep{qian2025dispider},
streaming models~\citep{yao2025timechatonline,zeng2025streamforest,zhang2026hermes,chen2026streamttt},
and the same-backbone, four-frame recent-window
SimpleStream baseline~\citep{shen2026simplestream}.
All are re-evaluated on the same question sets with visual-only inputs.
Frame budgets and sampling strategies remain model-specific, and computational budgets are not matched.

\textbf{Inference settings.}
SPARE samples at 1 FPS, anchored at the end of each interval.
The first pass reads up to four current frames. In the shared-ViT LoRA
configuration, the same LoRA-enabled model supplies the MLP gate's
prefill state and generates the answer.
Our evaluation implementation completes a current-only candidate answer.
When the gate triggers at threshold 0.5 and history is available,
Top-1 retrieval selects an earlier non-overlapping four-second clip;
a fresh second pass reads up to four historical and four current frames
chronologically and replaces the candidate with its response.
Otherwise, the first-pass response is returned.

\subsection{Benchmark Performance}
\label{sec:benchmark}
SPARE with shared ViT and LoRA achieves 77.78\% on OVO-Bench
(Table~\ref{tab:main_results}), outperforming SimpleStream by
10.17 percentage points and StreamTTT by 8.79.
Compared with SimpleStream, backward tracing improves from
53.99\% to 73.62\%, while real-time perception rises from
81.24\% to 81.93\%.
Hallucination detection shows the largest gain,
from 52.69\% to 90.32\%.
These results show improved backward tracing while preserving
real-time perception relative to SimpleStream.

All three SPARE variants outperform SimpleStream on
StreamingBench with visual-only inputs.
Shared ViT with LoRA achieves 73.31\%, exceeding SimpleStream
by 1.53 percentage points, with higher real-time visual and
contextual understanding scores but lower omni-source accuracy.
Without LoRA, native and shared ViT achieve 74.42\% and 74.22\%.
For the shared-ViT configuration, LoRA tuning and gate adaptation
increase the OVO-Bench score from 69.39\% to 77.78\%, while
the StreamingBench score decreases slightly from 74.22\% to 73.31\%.

Without LoRA, replacing the native retrieval encoder with
shared ViT changes OVO-Bench accuracy from 69.32\% to 69.39\%
and StreamingBench accuracy from 74.42\% to 74.22\%.
Encoder sharing thus maintains comparable overall performance
while removing the retrieval branch's standalone visual encoder.
\raggedbottom
\FloatBarrier
\section{Ablation Studies and Efficiency Analysis}
\label{sec:ablation}
\subsection{Ablation Studies}
\label{sec:ablation_frames}

\begin{table}[t!]
\centering
\small
\setlength{\tabcolsep}{6pt}
\caption{Retrieval policies with the same shared ViT, LoRA adapter,
and retrieval gate on OVO-Bench (\%). All is the mean
of the Real-time and Backward task-macro accuracies. History use is
the fraction of questions whose final responses use historical frames.}
\label{tab:retrieval_policy}
\begin{tabular*}{\linewidth}{@{\extracolsep{\fill}}lrrrr@{}}
\toprule
Policy & Real-time & Backward & All & History use \\
\midrule
Recent-only & 81.06 & 68.00 & 74.53 & 0.00 \\
Uniform history & 81.05 & 70.70 & 75.87 & 99.46 \\
Always retrieve & 81.16 & 73.22 & 77.19 & 99.46 \\
Gated retrieve & \textbf{81.93} & \textbf{73.62} & \textbf{77.78} & 40.74 \\
\bottomrule
\end{tabular*}
\end{table}

\setlength{\columnsep}{0.025\textwidth}
\makeatletter
\@afterindenttrue
\makeatother
\begingroup
\setlength{\intextsep}{0pt}
\begin{wraptable}{r}{0.48\textwidth}
    \centering
    \footnotesize
    \setlength{\tabcolsep}{4pt}
    \renewcommand{\arraystretch}{0.95}
    \setlength{\abovecaptionskip}{2pt}
    \setlength{\belowcaptionskip}{2pt}

    \caption{The impact of frames on OVO-Bench.}
    \label{tab:frame_ablation}

    \begin{tabular*}{\linewidth}
        {@{\extracolsep{\fill}}cccr@{}}
        \toprule
        \shortstack{Current\\frames} &
        \shortstack{Frames/\\clip} &
        \shortstack{Retrieved\\clips} &
        Score (\%) \\
        \midrule
        2 & -- & -- & 72.47 \\
        4 & -- & -- & 74.53 \\
        8 & -- & -- & 74.87 \\
        \midrule
        4 & 2  & 1 & 76.70 \\
        4 & 4  & 1 & \textbf{77.78} \\
        4 & 8  & 1 & 77.02 \\
        4 & 16 & 1 & 77.44 \\
        \midrule
        4 & 4 & 2, consecutive & 76.59 \\
        4 & 4 & 4, consecutive & 76.64 \\
        4 & 4 & 2, Top-2       & 77.28 \\
        4 & 4 & 4, Top-4       & 76.93 \\
        \bottomrule
    \end{tabular*}
\end{wraptable}

\noindent\textbf{Retrieval policy.}
Table~\ref{tab:retrieval_policy} fixes the trained components and
inference settings across four policies.
Recent-only uses up to four current frames.
Uniform history and Always retrieve add uniformly sampled history
and the Top-1 clip, respectively, with historical frame counts
matched per question at up to four.
Gated retrieve adds the Top-1 clip only when the gate triggers.

Uniform history improves Backward accuracy from 68.00\% to 70.70\%,
while Real-time accuracy remains nearly unchanged
(81.06\% versus 81.05\%).
With historical frame counts matched, Always retrieve further
improves Backward accuracy to 73.22\%, with 81.16\% on Real-time,
showing that selecting relevant clips primarily benefits
backward tracing.
Gated retrieve raises Real-time and Backward accuracy to 81.93\%
and 73.62\%, improving the overall score from 77.19\% to 77.78\%
while reducing history use from 99.46\% to 40.74\%.
Thus, selective retrieval improves both task groups while
using historical context less often.
Appendix~\ref{app:retrieval_policy_details} provides task scores,
the comparison protocol, and the native-encoder control without LoRA.

\noindent\textbf{Frame allocation and clip selection.}
Table~\ref{tab:frame_ablation} fixes the shared ViT, LoRA adapter,
and gate.
Increasing current frames from four to eight yields only
a small gain (74.53\% to 74.87\%).
With the same maximum of eight input frames, four current frames
plus one retrieved four-frame clip achieve 77.78\%, a gain of
2.91 percentage points over eight current frames.
Using eight or sixteen historical frames scores lower
(77.02\% and 77.44\%), favoring four recent and four historical
frames among the tested settings.

With four current frames and four frames per historical clip,
Top-2 outperforms two consecutive clips ending with the
highest-ranked clip (77.28\% versus 76.59\%); Top-4 likewise
outperforms four consecutive clips (76.93\% versus 76.64\%).
Both remain below single-clip retrieval (77.78\%).
Thus, relevance-based selection outperforms consecutive expansion,
but additional clips provide no accuracy gain in these comparisons.
\endgroup
\subsection{Efficiency Analysis}
\label{sec:efficiency_analysis}

\begin{figure}[!h]
    \centering
    \begin{minipage}[b]{0.49\linewidth}
        \centering
        \includegraphics[width=\linewidth]{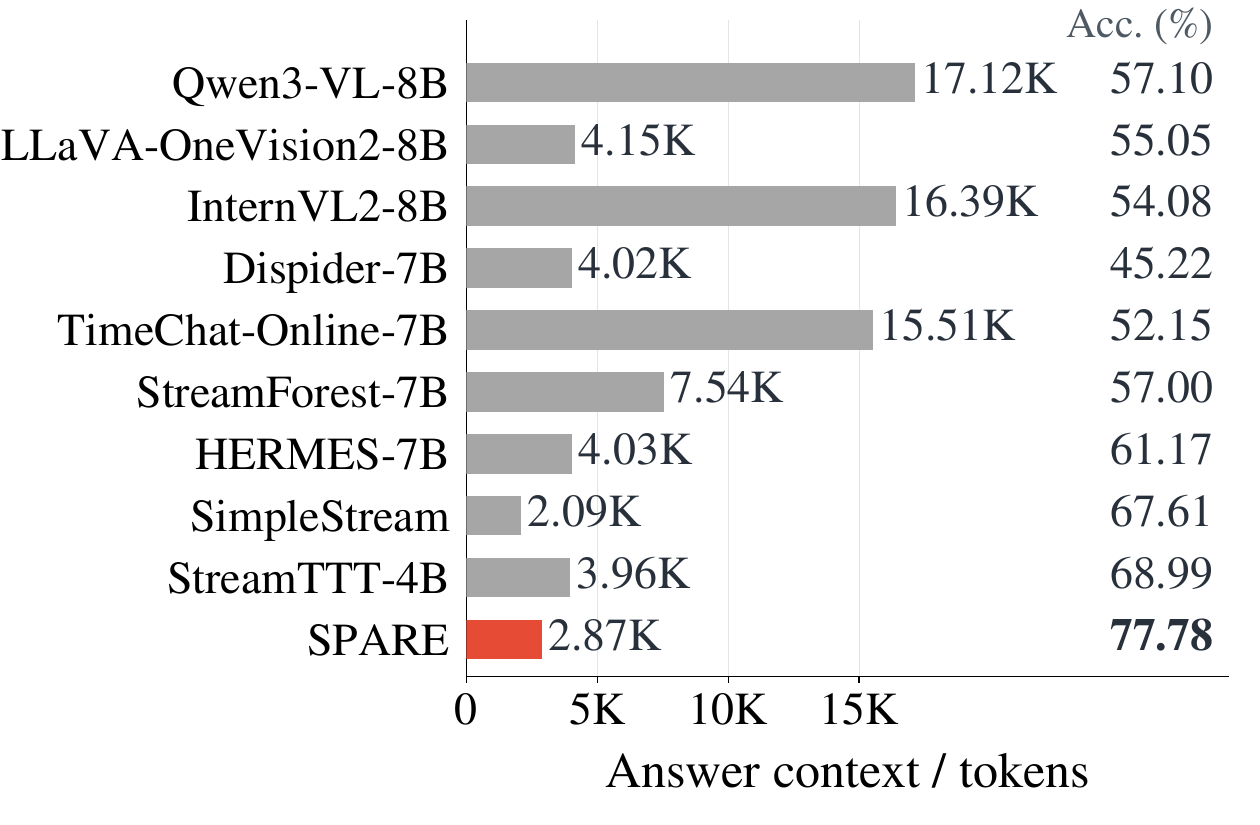}
        \par\smallskip
        {\small (a) Answering-context length}
    \end{minipage}\hfill
    \begin{minipage}[b]{0.49\linewidth}
        \centering
        \includegraphics[width=\linewidth]{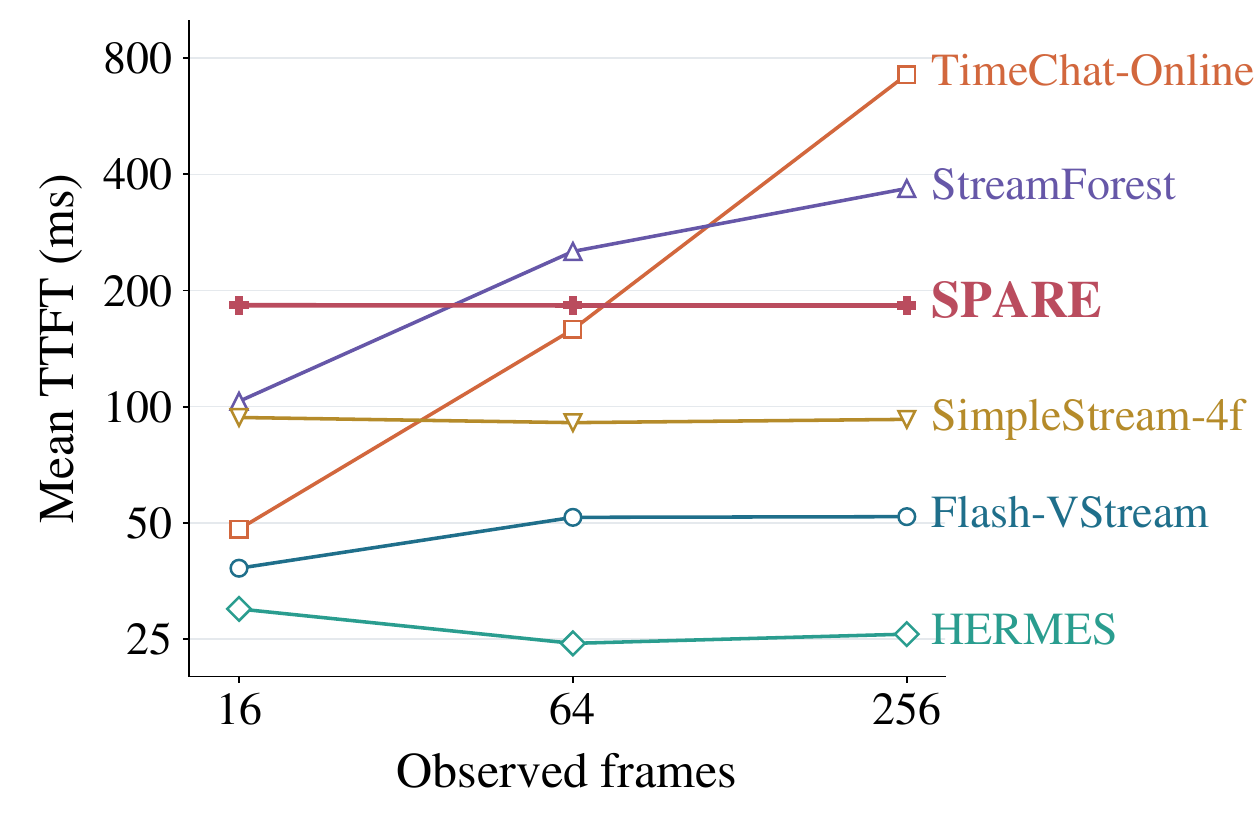}
        \par\smallskip
        {\small (b) Response latency}
    \end{minipage}
    \caption{Efficiency on OVO-Bench.
    \textbf{(a)} Mean per-question peak answering-context length and
    accuracy on all Real-time and Backward questions.
    \textbf{(b)} Mean time to first token on OVO-Bench's Real-time
    and Backward tasks at 16, 64, and 256 observed frames.}
    \label{fig:efficiency}
\end{figure}

\textbf{Answering-context length.}
SPARE uses a mean per-question peak answering context of
2.87K tokens, compared with 17.12K for Qwen3-VL-8B-Instruct
(Figure~\ref{fig:efficiency}(a)).
For each question, we take the maximum context length across passes,
then average over questions. Retrieved answering frames are included;
retrieval memory is excluded.
Relative to SimpleStream, SPARE uses an additional 0.78K tokens
(2.09K to 2.87K) while improving accuracy from 67.61\% to 77.78\%.

\textbf{Response latency.}
Figure~\ref{fig:efficiency}(b) reports mean time to first token
(TTFT) on OVO-Bench using one H100 80GB GPU and batch size one.
Timing starts at question arrival and ends when the first token
of the final response is available, including retrieval
and processing needed to produce that token but excluding
video preparation and state restoration.
For latency measurements, SPARE reuses cached visual features
and routes immediately after the gate's prefill decision,
skipping first-pass answer generation when retrieval is triggered.
Its mean TTFT remains near 183\,ms at 16, 64, and 256
observed frames.
Appendix~\ref{app:evaluation_details} provides the timing protocol.

\section{Conclusion}
We introduced \textsc{SPARE}, a streaming video understanding framework combining evidence sufficiency prediction, historical retrieval, and LoRA tuning for answering or abstention. The understanding and retrieval components can operate independently or share a visual encoder through four distilled linear adapters, removing the retrieval branch's separate visual encoder. Experiments on OVO-Bench and StreamingBench show improved accuracy over the base model and its recent-window baseline with visual-only inputs. 

\subsection*{AI use statement}
We used generative AI tools to assist with translation, language editing, literature search and reference formatting.
We also used Qwen3.8-27B~\citep{qwen2026qwen38} to assess whether video evidence was sufficient to answer questions and to filter training examples for LoRA instruction tuning and retrieval-gate training.
The data selection procedure retained examples only when two model assessments agreed, with additional checks on answer targets and multiple-choice options for LoRA instruction tuning, as detailed in Appendices~\ref{app:answer_tuning_details} and~\ref{app:alignment_gate_training}.
The authors take responsibility for the final manuscript, its claims, and the associated data and code, including content produced with AI assistance.

\bibliography{iclr2026_conference}
\bibliographystyle{iclr2027_conference}

\appendix

\section{Experimental Configuration}
\label{app:experimental_configuration}

\subsection{Shared-Encoder Alignment and Retrieval-Gate Training}
\label{app:alignment_gate_training}

\subsubsection{Shared visual encoder alignment}
\label{app:alignment_training}
We adapt the visual features of Qwen3-VL-8B-Instruct~\citep{bai2025qwen3vl} to the retrieval branch using the original Qwen3-VL-Embedding-2B~\citep{li2026qwen3vlembedding} as the teacher.
The distillation objective in Eq.~\ref{eq:distillation_loss} updates four $4608 \rightarrow 2048$ linear adapters, totaling 37.76M parameters.
All pretrained parameters remain frozen.
All student results in Table~\ref{tab:shared_retrieval} use joint
visual-token and final-embedding distillation.
For Eq.~\ref{eq:distillation_loss}, we fix
$(w_0,w_1,w_2,w_3)=(0.1,0.2,0.3,0.4)$ for the three DeepStack
branches and the final visual branch, respectively.
The visual cosine, visual SmoothL1, and final embedding cosine terms
have coefficients 1.0, 0.1, and 1.0, respectively.
These weights remain fixed throughout training.

Alignment uses 64K image entries from MMEB~\citep{jiang2025vlm2vec} and 64K video entries from LLaVA-Video-178K~\citep{zhang2025llavavideo}.
The image data are drawn equally from MSCOCO\_t2i and MSCOCO\_i2t~\citep{lin2014coco}, CIRR~\citep{liu2021cirr}, and NIGHTS~\citep{fu2023dreamsim}.
Videos come from the 0--30\,s duration bucket and are deduplicated by source video, then sampled at 1 FPS.
We alternate image and video batches and sample the four image subsets uniformly.

We train for 16,000 steps on 8 GPUs with a global batch size of 8, using BF16 mixed precision.
AdamW~\citep{loshchilov2019decoupled} uses learning rate $10^{-4}$ and weight decay 0.01.
After 500 steps of linear warmup, the learning rate follows a cosine schedule.
All subsequent experiments use the adapters saved at the final step.

\subsubsection{Initial retrieval-gate training}
\label{app:initial_gate_training}
The retrieval gate learns whether the current frames contain enough evidence to answer the question.
We train it on cached 4096-dimensional hidden states from the frozen Qwen3-VL-8B-Instruct.
Each state is taken from the last valid input token after the language model's final normalization, with LoRA and DeepStack disabled.
The gate has a hidden dimension of 256 and dropout 0.1, for approximately 1.06M trainable parameters.
Only the gate is updated.

The training set contains 15k automatically annotated questions
from multiple video datasets, each paired with 1--16 current
frames.
Qwen3.8-27B~\citep{qwen2026qwen38} assesses each question twice
using a prompt pair distinct from that used for LoRA annotation.
Both assessments receive identical current frames and timestamps,
without reference answers, answer choices, or each other's outputs.
A question is answerable only when the supplied frames establish
a concrete answer; contextual similarity or objects alone
do not establish an action.
We assign label 0 when both assessments find sufficient evidence
and label 1 when both find insufficient evidence.
Ambiguous or conflicting judgments are excluded.
Gate annotation uses the same model and generation settings as
LoRA annotation (Appendix~\ref{app:annotation_settings}), with
schema-constrained JSON decoding in both assessments.

We cache features for both open-ended and multiple-choice versions of each question.
Each epoch uses one version per question, split equally between 7,793 inputs of each format.
The gate is trained with the unweighted binary cross-entropy loss in Eq.~\ref{eq:gate_loss}.
We use AdamW with constant learning rate $10^{-3}$, weight decay 0.01, and batch size 128, running in FP32 on one H100 80GB GPU.
Subsequent experiments use the checkpoint saved after the first epoch.

\subsection{LoRA Instruction Tuning: Data and Implementation}
\label{app:answer_tuning_details}

\subsubsection{Data construction}
\paragraph{Candidate sources and hard negatives.}
All candidates come from audited video QA.
Training and validation splits keep source questions and videos separate, including videos linked by reuse of the same question.
A retrieved clip can be similar to a question without containing enough evidence to answer it.
We include such hard negatives in the data for LoRA instruction tuning alongside examples with sufficient evidence.
To find them, we use the frozen shared-encoder Embedding-2B retriever to rank previously audited unanswerable examples by the cosine similarity between each question and its Top-1 historical clip.
We prioritize the highest-scoring negatives, using the visual assessments to establish that the evidence is insufficient.

\paragraph{Retrieved evidence and answerability labels.}
We select 512 questions from different videos: 128 from current-answerable sources, 192 from history-assisted sources, and 192 from insufficient-evidence sources.
Within each source condition, we randomly shuffle the candidates and select equal numbers of open-ended and multiple-choice questions.
Multiple-choice questions retain their existing options.

We then assemble the visual input used for LoRA instruction tuning.
For each question, we keep the latest four-second window and retrieve the most similar earlier four-second clip using the native Qwen3-VL-Embedding-8B model~\citep{li2026qwen3vlembedding}.
This model selects the training clips offline, while the deployed SPARE model uses the shared-encoder Embedding-2B retriever.
The retrieval query includes the question and any answer options, and candidates are ranked by cosine similarity.
Both clips are sampled at 1 FPS from the start of their intervals, giving up to four historical frames followed by four current frames.
We preserve the timestamps, question, and options and apply native image preprocessing.

Answerability is assessed from this combined input.
Qwen3.8-27B~\citep{qwen2026qwen38} assesses the question
and timestamped history and current frames using two distinct
prompts.
Both assessments receive identical visual inputs, without
reference answers, answer options, or each other's outputs.
They require a concrete answer supported by visible evidence,
respect the question's temporal constraints, and reject
inferences about unseen events.
Examples are retained only when both assessments agree
on answerability; ambiguous or conflicting judgments are excluded.
For multiple-choice questions, a further visual check examines the options.
An answerable example must have exactly one supported option, whereas an unanswerable example must have no supported concrete answer.
These checks retain 427 candidates.
Because the label is assigned to the actual input, it can differ from the source label after retrieval and resampling.

\paragraph{Training-set selection.}
We select 48 examples for each combination of answerability and question format, giving 192 distinct examples.
Candidates are randomly shuffled within each group.
Each answerable group first includes up to 12 examples from current-answerable sources, so the model also learns to use current evidence when history is present.
We next review the existing answers from the two assessments
and the option-check results for target consistency.
This step uses no additional model call;
exclusions and final target selection are implemented in code.
This removes 17 answerable multiple-choice examples with conflicting or ambiguous option mappings, one open-ended example with conflicting action descriptions, and one unanswerable multiple-choice example with no abstention option.
All 173 remaining examples, each from a different video, are used for LoRA instruction tuning.

The final set contains 78 answerable and 95 unanswerable examples.
Among the multiple-choice questions, 31 are answerable and 47 are unanswerable.
The corresponding counts for open-ended questions are 47 and 48.
Targets use the reviewed answer or option when the evidence is sufficient, and the existing abstention option or an open-ended abstention otherwise.
The videos come from LLaVA-Video-178K~\citep{zhang2025llavavideo} (132),
DiDeMo~\citep{hendricks2017localizing} (19),
ActivityNet~\citep{cabaheilbron2015activitynet} (6),
QuerYD~\citep{oncescu2021queryd} (6),
COIN~\citep{tang2019coin} (5),
Charades~\citep{sigurdsson2016hollywood} (3),
and YouCook2~\citep{zhou2018procedures} (2).

\paragraph{Annotation model and decoding.}
\label{app:annotation_settings}
LoRA annotation uses the
\texttt{Qwen/Qwen3.8-27B-FP8} checkpoint, dequantized to BF16
for inference with SDPA attention and no additional adapters.
Generation uses greedy decoding
(\texttt{do\_sample=False}), a maximum of 512 new tokens,
and \texttt{enable\_thinking=False}.
The processor is loaded from the same checkpoint.
Schema constraints are applied only when rechecking malformed outputs.

\subsubsection{Training prompt construction}
\label{app:lora_training_prompts}
Each example is a separate conversation with three messages: a system instruction, a multimodal user input, and the target assistant response.
The user input presents the retrieved historical clip, the current window, and then the question.
No earlier model response or conversation state is carried over.

\paragraph{System instruction.}
Every example uses the following instruction:
\begin{quote}
\small
The input contains the current video clip and retrieved history, with timestamps. Answer using evidence relevant to the question's time or event, and follow its answer requirements.
\end{quote}

\paragraph{Visual context and timestamps.}
The user message begins with the query time.
Each clip is then introduced by its time interval, followed by its frames in chronological order.
We supply each frame as an image preceded by its timestamp.
All times are expressed in seconds to one decimal place.
In the template below, braces mark fields filled for each example, and image markers stand for actual image inputs.
Line breaks between frame items are shown for readability.
\begin{quote}
\small
\begin{verbatim}
The question is asked at video time {query_time} seconds.
{history_label}: [{history_start}, {history_end}) seconds
<{history_time_1} seconds> [historical image 1]
...
<{history_time_m} seconds> [historical image m]
Current segment: [{current_start}, {current_end}) seconds
<{current_time_1} seconds> [current image 1]
...
<{current_time_n} seconds> [current image n]
{question and answer-format instruction}
\end{verbatim}
\end{quote}
The historical clip keeps its retrieval label, such as \texttt{Retrieved segment 0}, and the current clip is labeled \texttt{Current segment}.
Intervals include their start time and exclude their end time.
The historical clip ends at or before the current window begins, and both clips lie within the video observed by the query time.
The retained examples contain one to four historical frames and four current frames.
The model receives image inputs as visual tokens, rather than the image markers shown in the template.

\paragraph{Question and answer format.}
The original question follows the frames.
For multiple-choice examples, options keep their stored order, including the position of any abstention option.
We label them consecutively from A and separate them with a semicolon and a space:
\begin{quote}
\small
\begin{verbatim}
{question}
Options: A. {option_0}; B. {option_1}; ...
Only give the best option's letter directly.
\end{verbatim}
\end{quote}
Open-ended questions use the following ending:
\begin{quote}
\small
\begin{verbatim}
{question}
Answer the question directly.
\end{verbatim}
\end{quote}
Answerability labels and audit explanations determine the training target but are not included in the prompt.

\subsubsection{Answer targets and loss masking}
For multiple-choice questions, the target is either the supported option letter or the existing abstention-option letter.
For open-ended questions, it is the reviewed short answer or \texttt{Unable to answer.}
The assistant produces the target directly, without an explanation or an \texttt{ANSWER} or \texttt{NO\_MATCH} prefix.
The end-of-turn token completes the supervised sequence: for example, \texttt{B<|im\_end|>} for option B or \texttt{Unable to answer.<|im\_end|>} for an open-ended abstention.

We format each conversation with the native Qwen3-VL chat template, using an empty tool list and disabling thinking.
The target is already present as the assistant message, so we do not append a generation prompt.
Frames are passed as individual images rather than as a video item.
The loss in Eq.~\ref{eq:sft_loss} applies to the answer text and its terminating \texttt{<|im\_end|>} token.
Every other position receives the ignore label $-100$: system and user text, visual tokens, role delimiters before the answer, padding, and the newline after the end token.

\subsubsection{Optimization and gate adaptation}
We initialize the understanding model from Qwen3-VL-8B-Instruct~\citep{bai2025qwen3vl} and add LoRA~\citep{hu2022lora} to the language model's query, key, value, and output projections, as well as its feed-forward gate, up, and down projections.
The adapters use rank 16, scaling parameter $\alpha=32$, and dropout 0.05, for 43.65M trainable parameters.
Only the LoRA parameters are updated.
The base model, visual encoder, retrieval branch, and retrieval gate remain frozen, and DeepStack is disabled in the answering branch.

Training visits each of the 173 examples once, using batch size 1 and no gradient accumulation.
We use AdamW~\citep{loshchilov2019decoupled} with initial learning rate $10^{-4}$, weight decay 0.01, and a gradient-norm limit of 1.0.
The learning rate follows cosine decay without warmup.
Training runs on one H100 80GB GPU in BF16 with FlashAttention-2~\citep{dao2024flashattention2} and non-reentrant gradient checkpointing, with TF32 disabled.
The maximum sequence length is 32,768 tokens, and every example fits without truncation.
We use the adapter saved after this single epoch for evaluation.

After LoRA instruction tuning, we adapt the retrieval gate to the LoRA-enabled model for the configuration in Table~\ref{tab:main_results}.
We freeze the LoRA adapter and extract new first-pass hidden states on the same 15k audited training questions used for the initial gate.
Starting from that gate's original checkpoint, we train for one epoch (122 steps) with unweighted binary cross-entropy.
AdamW uses learning rate $10^{-4}$, weight decay 0.01, batch size 128, and a gradient-norm limit of 1.0.
Each question appears in one format, giving 7,793 multiple-choice and 7,793 open-ended inputs.
At inference, LoRA is enabled in both answering passes, and the adapted gate triggers retrieval at threshold 0.5.

\subsection{Evaluation Details}
\label{app:evaluation_details}
OVO-Bench~\citep{li2025ovobench} contains 644 videos and 12 tasks covering current-scene perception, recall of past events, and response timing.
Its hallucination-detection task asks questions unsupported by the observed video.
The main comparison and Figure~\ref{fig:answer_context}a use the same
1,468 real-time and backward questions, excluding forward responding.
StreamingBench~\citep{lin2026streamingbench} contains 4,500 human-curated questions on 900 videos, with five query timestamps per video and 18 tasks across real-time visual, omni-source, and contextual understanding.

For the TTFT comparison in Figure~\ref{fig:efficiency}(b), we use
90 fixed questions, with 10 from each of the nine Real-time and
Backward tasks. Each question is measured three times at each of
16, 64, and 256 observed frames. We report mean TTFT across the
questions and repeats for each method and frame count.

\FloatBarrier

\section{Ablations and Supplementary Evaluations}
\label{app:ablation_evaluations}

\subsection{Choice of Retrieval Backbone}
\label{app:retriever_comparison}

We compare Qwen3-VL-Embedding-2B~\citep{li2026qwen3vlembedding} with CLIP ViT-L/14~\citep{radford2021learning}
and SigLIP2 SO400M Patch16 NaFlex~\citep{tschannen2025siglip2}
to evaluate their suitability for retrieving video
evidence. Qwen uses its pretrained visual encoder and language backbone,
without the shared encoder or alignment adapters used by SPARE.
All models are frozen and receive the same queries, candidate pools,
and source frames from the 18 MMEB-V2~\citep{meng2025vlm2vecv2}
video datasets (34,940 queries).
Qwen jointly encodes video and text, while CLIP and SigLIP2 average
normalized frame embeddings and combine text and visual embeddings
with equal weight for multimodal queries. Each model uses its native
image preprocessing and text limits. We rank candidates by cosine
similarity and report top-1 accuracy, averaging datasets equally.

Native Embedding-2B achieves the highest mean accuracy in all four
task families (Table~\ref{tab:mmeb_retriever_details}). Its overall
score is 56.13\%, compared with 33.38\% for CLIP and 33.93\% for
SigLIP2. Video and moment retrieval are especially relevant to SPARE
because they test whether a query retrieves the appropriate video or
temporal clip. On these tasks, native Embedding-2B scores 47.48\%
and 49.84\%, versus 32.78\% and 27.50\% for SigLIP2.
These results support choosing Qwen3-VL-Embedding-2B as the retrieval
backbone. The comparison evaluates pretrained components with different
model sizes and training histories.

\begin{table}[h]
\centering
\small
\caption{Top-1 accuracy on MMEB-V2 video tasks (\%). Each task-family
score averages its datasets equally; Overall averages all 18 datasets.
Qwen uses its native visual encoder. Bold marks the highest accuracy
in each row.}
\label{tab:mmeb_retriever_details}
\setlength{\tabcolsep}{5pt}
\renewcommand{\arraystretch}{1.05}
\begin{tabular*}{\linewidth}{@{\extracolsep{\fill}}lrrr@{}}
\toprule
Task family & CLIP & SigLIP2 & Native Embedding-2B \\
\midrule
Video classification & 44.98 & 38.11 & \textbf{64.21} \\
Video retrieval & 24.67 & 32.78 & \textbf{47.48} \\
Moment retrieval & 25.12 & 27.50 & \textbf{49.84} \\
Video QA & 35.47 & 34.76 & \textbf{60.47} \\
\midrule
Overall & 33.38 & 33.93 & \textbf{56.13} \\
\bottomrule
\end{tabular*}
\end{table}

\subsection{Effect of Encoder Sharing on Retrieval}
\label{app:shared_retrieval_comparison}
Using the retrieval backbone selected in Appendix~\ref{app:retriever_comparison},
we compare the native Embedding-2B teacher with the shared-encoder
student trained using the joint objective in Eq.~\ref{eq:distillation_loss}.
The student replaces the teacher's visual tower with the frozen
8B-Instruct encoder and the four adapters saved at step 16,000;
both use the same frozen Embedding-2B language backbone and text path.
We hold the 34,940 queries, candidate pools, source frames,
instructions, preprocessing, and ranking procedure fixed.
Query and candidate embeddings are computed separately for each model,
with last-valid-token pooling and $L_2$ normalization.
These evaluations rank candidates directly, without the answering
LoRA adapter or retrieval gate. Appendix~\ref{app:alignment_training}
provides the student's alignment training settings.

\begin{table}[ht]
\centering
\small
\caption{Direct retrieval with native teacher and shared student on
MMEB-V2. Accuracy and success rates are percentages; MRR is multiplied
by 100. Each column is an unweighted mean over the indicated datasets.
The five text-to-video datasets are DiDeMo, MSR-VTT, MSVD, VATEX,
and YouCook2.}
\label{tab:shared_retrieval}
\setlength{\tabcolsep}{5pt}
\renewcommand{\arraystretch}{1.05}
\begin{tabular*}{\linewidth}{@{\extracolsep{\fill}}lrrrr@{}}
\toprule
& \multicolumn{2}{c}{All 18 datasets} & \multicolumn{2}{c}{Text-to-video (5 datasets)} \\
\cmidrule(lr){2-3}\cmidrule(lr){4-5}
Metric & Native teacher & Shared student & Native teacher & Shared student \\
\midrule
Top-1 accuracy & 56.13 & 51.90 & 47.48 & 43.40 \\
Top-5 success rate & 87.25 & 84.83 & 72.09 & 68.09 \\
Top-10 success rate & 93.25 & 91.84 & 80.08 & 76.55 \\
MRR $\times 100$ & 69.49 & 66.06 & 58.74 & 54.77 \\
\bottomrule
\end{tabular*}
\end{table}

Table~\ref{tab:shared_retrieval} reports top-1 retrieval accuracy,
top-5 and top-10 retrieval success rates, and mean reciprocal rank (MRR).
Top-1 accuracy is the percentage of queries whose highest-ranked
candidate is relevant. Top-5 and top-10 success rates count queries
with at least one relevant candidate among the first five or ten results.
MRR averages the reciprocal of the first relevant candidate's rank,
assigning zero when no relevant candidate is retrieved.
We average each metric equally across datasets.
Alongside the full suite, we report the five text-to-video retrieval
datasets: DiDeMo, MSR-VTT, MSVD, VATEX, and YouCook2.
This separates video retrieval from the classification, moment
retrieval, and video QA tasks included in the overall score.

Table~\ref{tab:shared_retrieval} shows that, across the full suite,
sharing reduces top-1 accuracy by 4.22 percentage
points and the top-10 success rate by 1.41 points. The student's top-1
accuracy is 92.47\% of the teacher's score. On text-to-video retrieval,
top-1 accuracy decreases from
47.48\% to 43.40\%. The shared encoder therefore retains most of the
teacher's retrieval performance with a measurable loss in ranking quality.
Together with the downstream comparisons in Section~\ref{sec:benchmark},
these results quantify the trade-off between removing a separate visual
tower and preserving direct retrieval accuracy.

\subsection{Detailed Retrieval-Policy Results}
\label{app:retrieval_policy_details}

\subsubsection{Shared encoder with fixed LoRA supervision}
Both answering passes use the same LoRA checkpoint and user-only
prompt template, with answering DeepStack disabled and retrieval
DeepStack enabled. End-anchored sampling uses 1 FPS; eligible historical
clips end before the current window begins. All four policies retain
the current frames, spatial preprocessing, and greedy decoding.

Uniform history samples uniformly spaced frames from the pooled unique
frames of all eligible clips, matching the Top-1 clip's actual frame
count per question. It computes the ranking for budget matching, so
this comparison measures accuracy rather than deployment cost.
History is available for 1,460 of 1,468 questions; the fixed gate uses
it for 598 at threshold 0.5. First-pass answers agree across all four
policies, as do second-pass answers for gated and always retrieval
where both use history.

Table~\ref{tab:retrieval_policy_tasks} gives all nine task scores,
including the LoRA-enabled recent-only setting in
Table~\ref{tab:frame_ablation}. Gating improves EPM and ASI while
retaining the recent-only HLD score of 90.32\%.
The Backward and Real-time macro averages improve by 5.62 and 0.87
percentage points, respectively.

\begin{table}[!htbp]
\centering
\small
\caption{Task-level accuracy for the shared-ViT, LoRA-enabled OVO-Bench
policies (\%). The first six tasks form Real-time and the final three
form Backward. All weights the two task-group macro averages equally.
Bold marks the highest score in each row, including ties.}
\label{tab:retrieval_policy_tasks}
\setlength{\tabcolsep}{4pt}
\begin{tabular*}{\linewidth}{@{\extracolsep{\fill}}lrrrr@{}}
\toprule
Task & Recent-only & Uniform history & Always retrieve & Gated retrieve \\
\midrule
OCR & 93.96 & 93.96 & 93.96 & \textbf{95.97} \\
ACR & 84.40 & 83.49 & 84.40 & \textbf{85.32} \\
ATR & 81.90 & 79.31 & \textbf{83.62} & \textbf{83.62} \\
STU & \textbf{70.79} & \textbf{70.79} & 68.54 & \textbf{70.79} \\
FPD & 73.27 & \textbf{77.23} & 73.27 & 73.27 \\
OJR & 82.07 & 81.52 & \textbf{83.15} & 82.61 \\
\midrule
EPM & 52.19 & 56.23 & \textbf{60.94} & 60.27 \\
ASI & 61.49 & 65.54 & 68.92 & \textbf{70.27} \\
HLD & \textbf{90.32} & \textbf{90.32} & 89.78 & \textbf{90.32} \\
\midrule
Real-time macro & 81.06 & 81.05 & 81.16 & \textbf{81.93} \\
Backward macro & 68.00 & 70.70 & 73.22 & \textbf{73.62} \\
All & 74.53 & 75.87 & 77.19 & \textbf{77.78} \\
\bottomrule
\end{tabular*}
\end{table}

\subsubsection{Native-encoder control without LoRA}
Table~\ref{tab:native_retrieval_policy} retains the native-ViT,
no-LoRA control from Table~\ref{tab:main_results}, holding its model,
prompts, preprocessing, sampling, and initial gate fixed.
Gating reduces history use from 99.46\% to 39.51\% with a 0.32-point
accuracy decrease. This comparison isolates policy effects within
the native configuration, whose encoder, answering adapter, gate,
and second-pass prompt differ from the shared-ViT LoRA configuration.

\begin{table}[!htbp]
\centering
\small
\caption{Retrieval-policy control with native visual encoders and no
LoRA on the same 1,468 OVO-Bench questions (\%). History use counts
questions whose final responses use historical frames.}
\label{tab:native_retrieval_policy}
\setlength{\tabcolsep}{6pt}
\begin{tabular*}{\linewidth}{@{\extracolsep{\fill}}lrrrr@{}}
\toprule
Policy & Real-time & Backward & All & History use \\
\midrule
Recent-only & 81.11 & 50.39 & 65.75 & 0.00 \\
Always retrieve & 82.34 & \textbf{56.95} & \textbf{69.64} & 99.46 \\
Gated retrieve & \textbf{82.37} & 56.28 & 69.32 & 39.51 \\
\bottomrule
\end{tabular*}
\end{table}
\FloatBarrier

\section{Example Questions and Model Responses}
\label{app:qualitative_examples}
\raggedbottom
We present five examples from the shared-ViT SPARE model with LoRA instruction tuning and the adapted MLP retrieval gate.
The OVO-Bench~\citep{li2025ovobench} examples show two correct answers and a correct final abstention.
The MMVU~\citep{zhao2025mmvu} examples show a direct correct answer and an initial abstention followed by a correct answer after retrieval.
All five final responses match the dataset references.
Each row shows the four frames actually supplied from a clip, with timestamps in seconds relative to the source video.

\newcommand{\exampleframes}[6]{%
  \par\smallskip\noindent{\small\textbf{#1}}\par\nobreak
  \noindent\includegraphics[width=\linewidth]{#2}\par\nobreak
  \noindent\makebox[0.25\linewidth]{\scriptsize #3 s}%
  \makebox[0.25\linewidth]{\scriptsize #4 s}%
  \makebox[0.25\linewidth]{\scriptsize #5 s}%
  \makebox[0.25\linewidth]{\scriptsize #6 s}\par
}

\par\medskip\noindent
\begin{minipage}{\linewidth}
\textbf{Multiple-choice answer: OVO-Bench, EPM 108.}\par\nobreak\smallskip\noindent
\textbf{Question:} What did I put in the container?\\
\textbf{Options:} A. cement; B. water; C. mud; D. garbage.\\
\textbf{Model response:} \texttt{A}\\
\textbf{Reference answer:} A. cement.
\exampleframes{Retrieved clip: $[188,192)$ s}
  {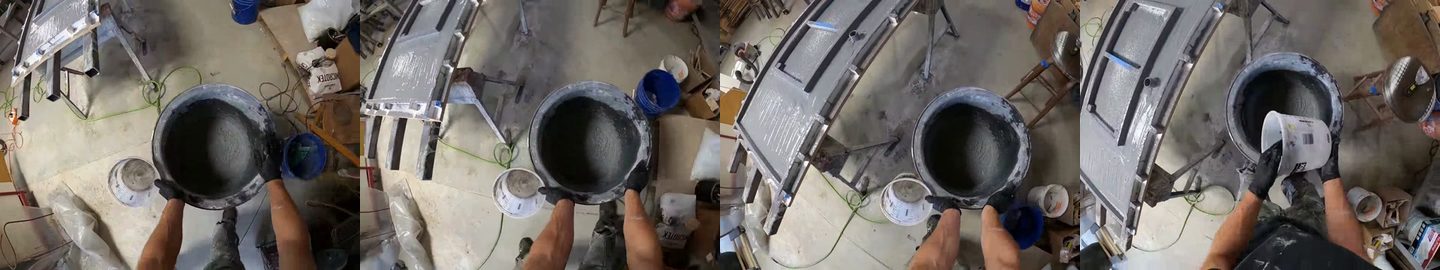}
  {189.000}{190.000}{191.000}{191.967}
\exampleframes{Current clip: $[317.95,321.95)$ s}
  {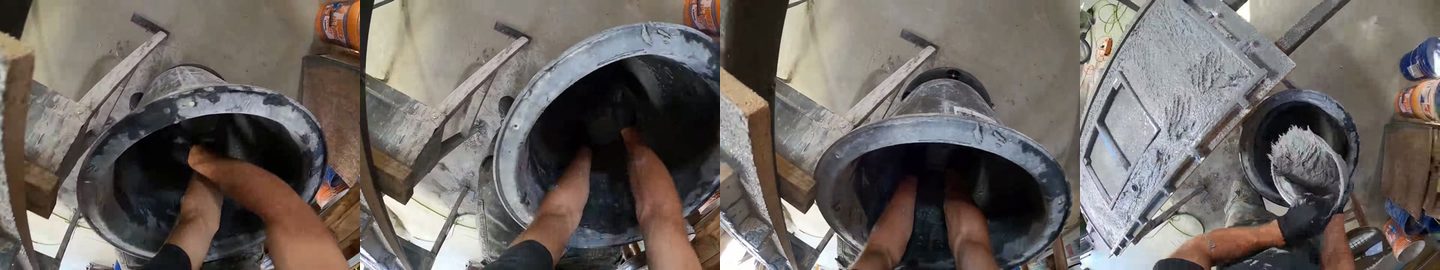}
  {318.933}{319.933}{320.933}{321.933}
\end{minipage}

\par\medskip\noindent
\begin{minipage}{\linewidth}
\textbf{Multiple-choice correction after retrieval: OVO-Bench, ASI 599.}\par\nobreak\smallskip\noindent
\textbf{Question:} Which object did the person close before they took the cup/glass/bottle?\\
\textbf{Options:} A. The door.; B. The refrigerator.; C. The closet/cabinet.; D. The paper/notebook.\\
\textbf{Current-only response:} \texttt{C}\\
\textbf{Final response after retrieval:} \texttt{B}\\
\textbf{Reference answer:} B. The refrigerator.
The current-only response selects the closet/cabinet. After retrieval, the response changes to the refrigerator, matching the reference.
\exampleframes{Retrieved clip: $[0,4)$ s}
  {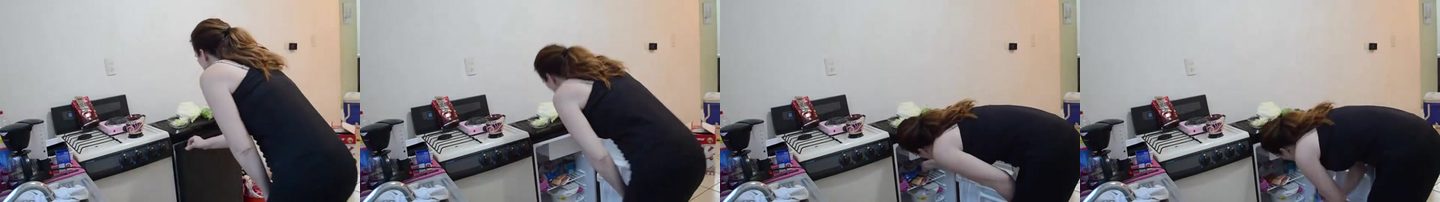}
  {1.000}{2.000}{3.000}{3.958}
\exampleframes{Current clip: $[16.954,20.954)$ s}
  {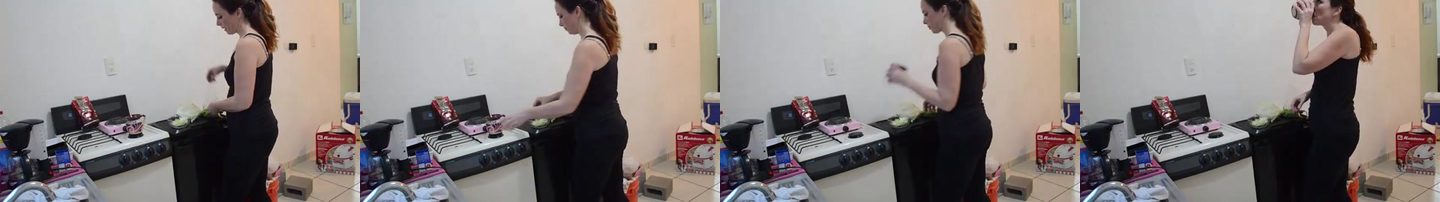}
  {17.917}{18.917}{19.917}{20.917}
\end{minipage}

\par\medskip\noindent
\begin{minipage}{\linewidth}
\textbf{Multiple-choice abstention: OVO-Bench, HLD 298.}\par\nobreak\smallskip\noindent
\textbf{Question:} Where did I put the vacuum cleaner head?\\
\textbf{Options:} A. closet; B. Unable to answer; C. bathroom; D. kitchen.\\
\textbf{Model response:} \texttt{B}\\
\textbf{Reference answer:} B. Unable to answer.
The model selects the abstention option even after consulting retrieved history.
\exampleframes{Retrieved clip: $[216,220)$ s}
  {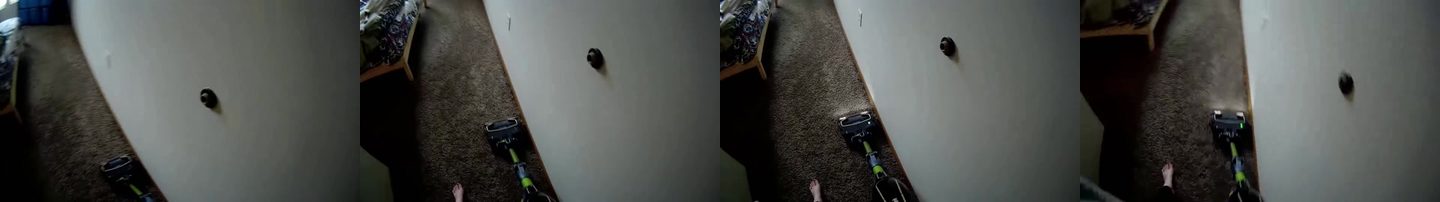}
  {217.000}{218.000}{219.000}{219.967}
\exampleframes{Current clip: $[387.95,391.95)$ s}
  {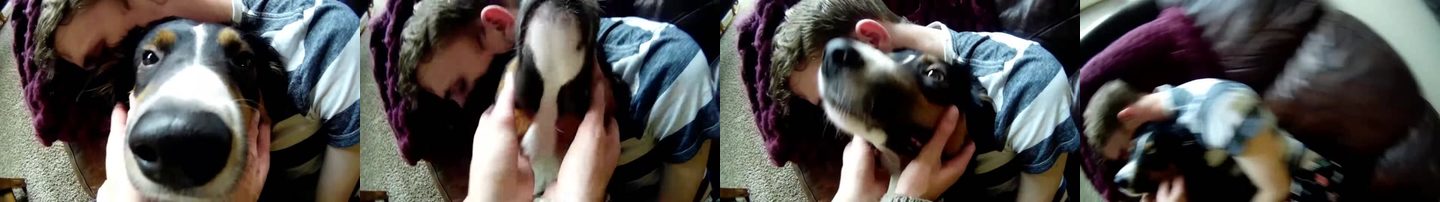}
  {388.933}{389.933}{390.933}{391.933}
\end{minipage}

\par\medskip\noindent
\begin{minipage}{\linewidth}
\textbf{Open-ended answer: MMVU validation 22.}\par\nobreak\smallskip\noindent
\textbf{Question:} The video demonstrates a famous experiment that proves the existence of the \_\_\_\_\_.\\
\textbf{Model response:} \texttt{blind spot}\\
\textbf{Reference answer:} blind spot.
The model answers from the current window without retrieval.
\exampleframes{Current clip: $[13.017,17.017)$ s}
  {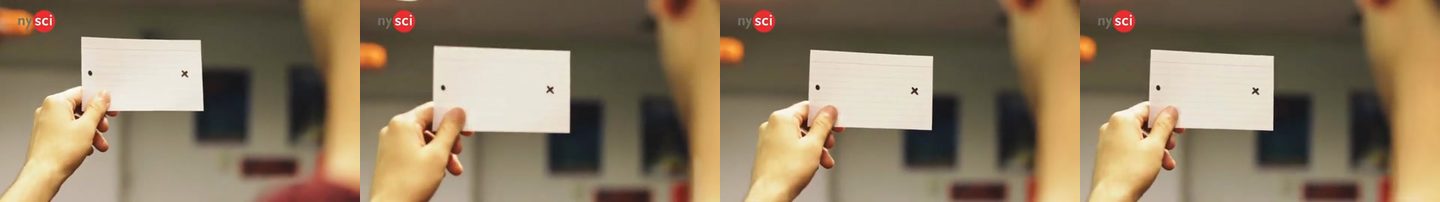}
  {14.014}{15.015}{16.016}{16.984}
\smallskip\textit{No retrieved clip: the gate did not trigger retrieval.}
\end{minipage}

\par\bigskip\noindent
\begin{minipage}{\linewidth}
\textbf{Open-ended initial abstention: MMVU validation 6.}\par\nobreak\smallskip\noindent
\textbf{Question:} What is the algorithm shown in the video?\\
\textbf{Current-only response:} \texttt{Unable to answer.}\\
\textbf{Final response after retrieval:} \texttt{Binary Search}\\
\textbf{Reference answer:} Binary Search.
With the retrieved clip added, the model replaces its initial abstention with the correct answer.
\exampleframes{Retrieved clip: $[4,8)$ s}
  {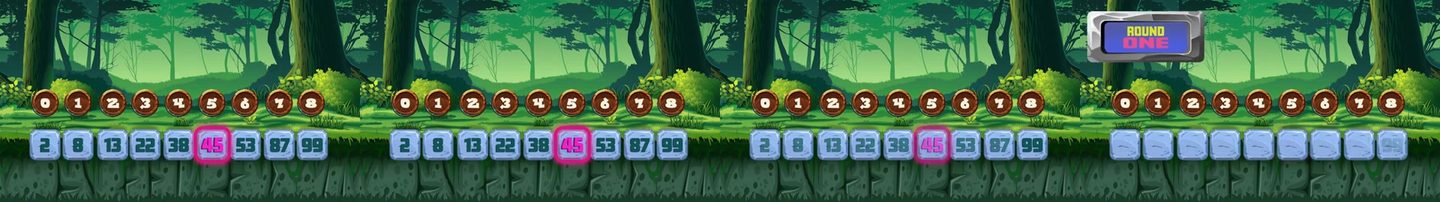}
  {5.000}{6.000}{7.000}{7.960}
\exampleframes{Current clip: $[96,100)$ s}
  {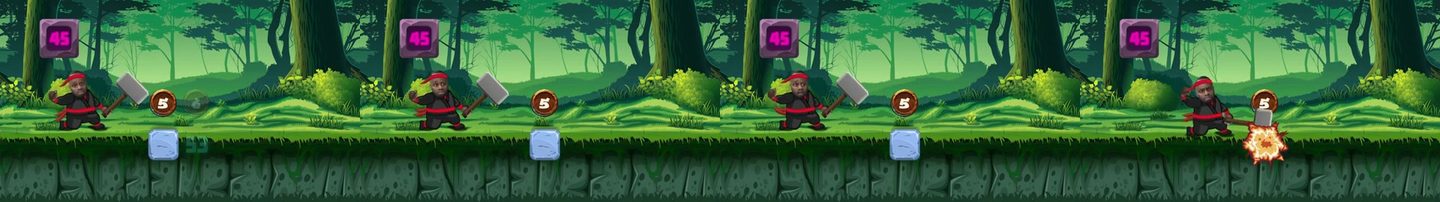}
  {97.000}{98.000}{99.000}{99.960}
\end{minipage}

\end{document}